\pdfoutput=1

\documentclass[11pt]{article}

\usepackage[final]{acl}
\usepackage{times}
\usepackage{latexsym}
\usepackage[T1]{fontenc}

\usepackage[utf8]{inputenc}
\usepackage{microtype}
\usepackage{inconsolata}
\usepackage{graphicx}

\usepackage[inline]{enumitem}

\usepackage{amsmath}
\usepackage{float}
\usepackage{booktabs}
\usepackage{multirow}

\usepackage{soul}
\usepackage{tabularx}

\newcommand{\heading}[1]{\vspace*{.5mm}\noindent\textbf{#1.}}

\title{Controlling Risk of Retrieval-augmented Generation: \\ A Counterfactual Prompting Framework}

\author{Lu Chen\textsuperscript{1,2}, 
Ruqing Zhang\textsuperscript{1,2}, 
Jiafeng Guo\textsuperscript{1,2}\thanks{Corresponding author}, 
Yixing Fan\textsuperscript{1,2}, 
Xueqi Cheng\textsuperscript{1,2}\\
\textsuperscript{1}CAS Key Lab of Network Data Science and Technology, ICT, CAS, Beijing, China\\
\textsuperscript{2}University of Chinese Academy of Sciences, Beijing, China\\
\{chenlu19z, zhangruqing, guojiafeng, fanyixing, cxq\}@ict.ac.cn
 \\
}

\begin{document}
\maketitle
\begin{abstract}

Retrieval-augmented generation (RAG) has emerged as a popular solution to mitigate the hallucination issues of large language models. 
However, existing studies on RAG seldom address the issue of predictive uncertainty, i.e., how likely it is that a RAG model's prediction is incorrect, resulting in uncontrollable risks in real-world applications. 
In this work, we emphasize the importance of risk control, ensuring that RAG models proactively refuse to answer questions with low confidence. 
Our research identifies two critical latent factors affecting RAG's confidence in its predictions: the quality of the retrieved results and the manner in which these results are utilized.
To guide RAG models in assessing their own confidence based on these two latent factors, we develop a counterfactual prompting framework that induces the models to alter these factors and analyzes the effect on their answers.
We also introduce a benchmarking procedure to collect answers with the option to abstain, facilitating a series of experiments. 
For evaluation, we introduce several risk-related metrics and 
the experimental results demonstrate the effectiveness
of our approach. 
Our code and benchmark dataset are available at https://github.com/ict-bigdatalab/RC-RAG.

\end{abstract}

\section{Introduction}

\begin{figure}[t]
    \centering
    \includegraphics[width=\linewidth]{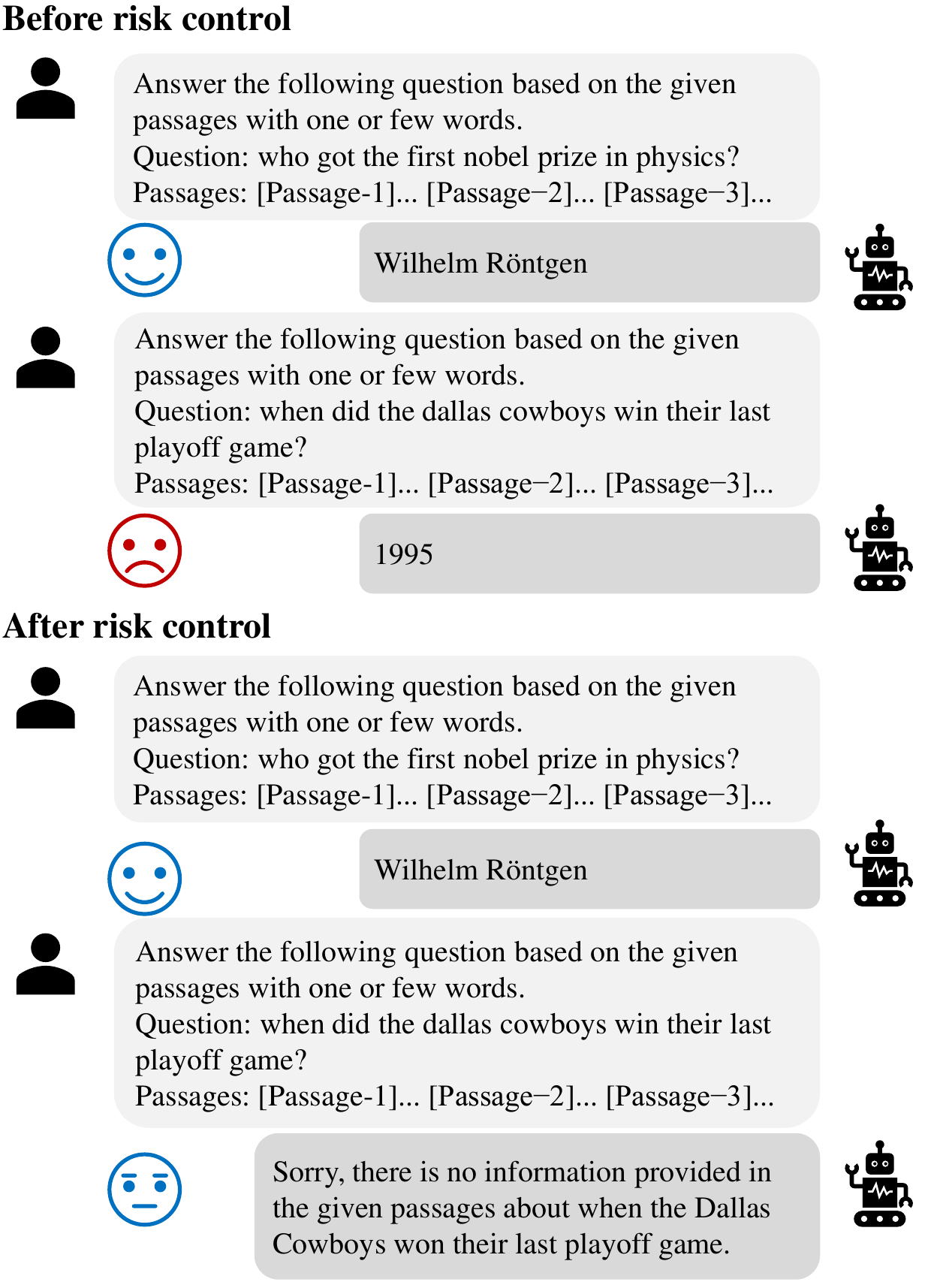}
    \vspace{-4mm}
    \caption{Illustration of risk control for RAG. Given a  question, a risk controlled RAG model is expected to provide the correct answer if it has knowledge of the question, or alternatively, refuses to answer the question.}
    \label{fig:intro}
    \vspace{-2mm}
\end{figure}

Large language models (LLMs) have gained considerable attention across a wide range of language tasks \cite{brown2020language,kandpal2023large,li2023blip,touvron2023llama}. 
Despite the exciting performance, LLMs may suffer from hallucination issues~\cite{ye2023cognitive,azamfirei2023large}, due to limited memorization abilities or outdated pre-training corpora  \cite{longpre2021entity,xie2023adaptive}.
Recently, retrieval-augmented generation (RAG) has emerged as a promising solution to enhance factual accuracy \cite{kandpal2023large,xie2023adaptive,gao2023retrieval}, by synthesizing text snippets retrieved from external resources into final responses  \cite{zhu2023large,ram2023context,izacard2023atlas,petroni2021kilt,ai2023information}. 

However, directly applying existing RAG techniques, particularly for knowledge-intensive tasks \cite{thorne2018fever,yang2018hotpotqa,petroni2021kilt} such as factoid question answering  \cite{aghaebrahimian2016open,aghaebrahimian2018linguistically}, introduces significant risks in practice.
When confronted with noisy search results, even the most advanced RAG models are prone to producing unreliable answers, often exhibiting overconfidence in these erroneous responses \cite{yang2023alignment,ren2023investigating}. 
Such unreliable answers may severely undermine the user's question answering (QA) experience.
Therefore, for practical applications, especially in sensitive domains like healthcare and legal assistance, it is crucial that RAG systems confidently provide answers when they know and state ``I don't know'' when they do not, as illustrated in Figure \ref{fig:intro}. 
This calls for the investigation on the risk control issue of RAG, a core research problem we want to tackle in this work. 
This approach reflects wisdom, as it involves RAG models proactively refusing to answer questions when predictions are uncertain.

Unfortunately, most previous research on risk control has focused on LLMs \cite{tian2023just,lin2023generating,feng2024don}. 
There has been little work addressing the predictive uncertainty issue of RAG. 
Compared to the uncertainty assessment of LLMs, which concentrates on internal knowledge boundaries, the assessment for RAG requires additional consideration of external knowledge from retrieved results.
In this work, we identify two critical factors during the uncertainty assessment of RAG: \emph{the quality of the retrieved results and the manner in which they are used}. 
This raises an important research question: \emph{how can we assess the predictive uncertainty of RAG based on these two retrieval results-related factors to determine when to discard or keep the generated answers}?

In this work, we propose a new task of risk control for RAG (RC-RAG) to decide whether to keep or discard the RAG outputs based on confidence assessment. 
We then introduce a novel \emph{counterfactual prompting framework for RAG} under the zero-shot scenario, leveraging the counterfactual thinking for confidence assessment based on two latent factors. 
Counterfactual \cite{pearl2009causality} describes the human capacity to learn from prior experiences by imagining the outcomes of alternative actions that could have been taken. 
For a language model, we can inject counterfactual thinking into prompt, like ``what if...'' or ``assume that'', to imagine or simulate the consequences of changing a factor.
Here, we induce the model to imagine scenarios where the quality of the retrieved results and their usage are poor, then measure its confidence based on the effect of these imagined scenarios on the answers.
Specifically, our framework consists of three major modules, i.e., a prompting generation module, a judgment module, and a fusion module:
\begin{enumerate*}[label=(\roman*)]
\item the prompting generation module generates answers under two scenarios that challenges the improper use and poor quality of the retrieved results, respectively;
\item the judgment module determines whether to discard or keep the generated answers for both scenarios; and 
\item the fusion module combines the judgment results from both scenarios to produce the final decision for selective output.
\end{enumerate*}
It is important to note that our method is a general post-processing technique, making it applicable to almost any existing RAG method.

For evaluation, traditional metrics like Exact Match and F1 score typically focus on the effectiveness of RAG. In this work, we propose four risk-related metrics - \emph{risk}, \emph{carefulness}, \emph{alignment}, and \emph{coverage} - for risk-aware RAG evaluation. 
Due to the limited availability of datasets directly applicable to RC-RAG, we have constructed a novel risk control benchmark based on two publicly available QA datasets. 
Extensive experiments on RAG with Mistral \cite{jiang2024mixtral} and ChatGPT \cite{roumeliotis2023chatgpt} as backbones demonstrate that the proposed framework can effectively abstain, outperforming baselines in 3 out of the 4 settings in terms of carefulness and risk, with up to a 14.76\% improvement in carefulness and a 2.88\%  reduction in risk on average.

\vspace{-2mm}
\section{Related work}
\heading{Retrieval-augmented generation}
The typical retrieval-augmented generation (RAG) method follows a retrieve-then-generate pipeline, first retrieving relevant documents from a grounding corpus and then generating the final answer by the frozen generators \cite{shi2023replug,ram2023context}.
The retrieval augmentation is performed for all the questions through a single round \cite{lewis2020retrieval,guu2020realm,izacard2021leveraging,shi2023replug} or multiple rounds \cite{borgeaud2021improving,ram2023context,trivedi2023interleaving,jiang2023active,liu2024ra}.
However, such practice sometimes hurt generation performance, due to the unsatisfactory retrieved results \cite{mallen2023not,ren2023investigating,yoran2023making,tan2024blinded}.
The reason may lie in the inconsistency between the relevance judgments in retrieval stage and the utility judgments in generation stage \cite{zhang2024large}.
Besides jointly optimization of the retriever and generator \cite{guu2020realm,lewis2020retrieval,singh2021end,izacard2023atlas}, another solution is adaptive retrieval augmentation \cite{jiang2023active,asai2023self,wang2023self}, which actively determines when to retrieve based on internal knowledge boundaries. 

\heading{Knowledge boundary}
Detecting what LLMs know and do not know measures the boundary of models’ internal knowledge, which can be applied to determine when to abstain it \cite{kadavath2022language,yang2023alignment}.
The basic realization involves prompting one LLM to either verify in advance or to self-reflect on its response afterward \cite{ren2023investigating,li2024confidence}.
It works for almost all LLMs, but there is a problem of overconfidence \cite{yin2023large}.
Self-consistency between multiple inference also reflects the models' answering ability \cite{manakul2023selfcheckgpt}, which is widely applicable but of high cost.
Calibration-based methods obtain uncertainty or confidence scores of answers based on factors such as entropy, and token probability \cite{lin2023generating,yang2023alignment}. A threshold is set to reject answers with low scores.
Besides, some work elicits self-knowledge by referring to existing cases, which needs labeled samples.
Through instruction tuning \cite{ouyang2022training} or applying a small trainable model as classifier \cite{slobodkin2023curious,azaria2023internal}, LLMs can choose to abstain the answer when facing new questions.
However, the limitation of the aforementioned work is that it only examines confidence when using internal knowledge, without considering the confidence when integrating external knowledge under the RAG setting.
Though some work deals with knowledge conflict between the internal knowledge and external knowledge \cite{li2023large,xie2023adaptive,qian2023merge,tan2024blinded}, it seldom rejects the RAG results, under the assumption that at least one kind of knowledge is true.
This assumption is not conducive to risk control of RAG, since the retrieval results may contain noise.
Therefore, in this work, we explore possible ways to control risk by discarding the RAG results, especially designed for external knowledge from retrieval results.

\heading{Counterfactual thinking}
As the third level of the causal ladder after association and intervention, counterfactual reflects causality by imagining “what would the outcome be had the variable(s) been different” \cite{pearl2009causality,nan2021uncovering}.
Counterfactual inference helps model unchanging causal mechanisms for better generalization and debias, which can be utilized for text classification, visual question answering, recommendation system and so on \cite{qian2021counterfactual,niu2021counterfactual,wei2021model,wang2022causal,deng2023counterfactual}.
It can calibrate causal effects through mediation analysis, by estimating the total effect and then eliminating the undesired effect \cite{xie2021factual}.
Different from these works, we focus on injecting counterfactual thinking into the prompt to better apply retrieval-augmented LLMs.

\vspace{-2mm}
\section{Problem statement}

\subsection{Task description}
The RC-RAG task aims at assessing confidence or uncertainty of RAG answer to enable risk control in RAG.
Formally, given a question $Q$ and a group of retrieved passages $P$, the task outputs the answer $A$ along with a judgment label $J\in\{0,1\}$.
For the samples with high confidence, the judgment label $J$ is set as 1, indicating that the RAG answer could be \textit{kept}.
Oppositely, $J=0$ is set for those uncertain output of RAG, which should be \textit{discarded}.
Ideally, the assessment of confidence should align with the extent to which RAG knowledge supports the correct answer.

\subsection{Benchmark}

\heading{Data}
To our best knowledge, there is limited available dataset that can be directly used for risk control for RAG. Therefore, we construct a RC-RAG benchmark composed of quadruple <$Q,P,A,J$> through automatic annotation.
In the following, we introduce the data source and collection process of this benchmark.

\noindent\emph{Data source.}
In this work, we focus on factoid question answering (FQA) \cite{aghaebrahimian2016open,aghaebrahimian2018linguistically}, which typically provides a limited number of short answers, such as entities or numbers, and therefore carries a higher risk compared to non-factoid QA.
We collect questions from two widely used datasets including Natural Questions (NQ) \cite{kwiatkowski2019natural} and TriviaQA (TQ) \cite{joshi2017triviaqa}.
Since we focus on a zero-shot scenario, we collect question $Q$ from their test sets.

\noindent\emph{Data collection.}
We further collect $P,A,J$ based on questions $Q$ in the data source.
\begin{itemize}[leftmargin=*]
\item \textbf{Passage collection.} For each question $q\in Q$, we utilize a dense retriever to retrieve top-k relevant passages $p=\{p_{1},...,p_{k}\}$ from external resources.

\begin{table}[t]
    \centering
   \renewcommand{\arraystretch}{1}
    \setlength\tabcolsep{7pt}
    \begin{tabular}{ccccc}
        \toprule
         & \multicolumn{2}{c}{\textbf{RC-TQ} (7785)}   & \multicolumn{2}{c}{\textbf{RC-NQ} (3610)}  \\
        \cmidrule(r){2-3}
        \cmidrule(r){4-5}
        & TQ-A& TQ-U& NQ-A& NQ-U \\
        \midrule
        ChatGPT & 5551 & 2234 & 1785 & 1825\\
        Mistral & 5553 & 2232 & 1830 & 1780\\
        \bottomrule
    \end{tabular}
    \caption{Statistics of the full test sets and annotated results of answerable (A) and unanswerable (U) samples.}
    \label{tab:statistics}
    \vspace{-4mm}
\end{table}

\item \textbf{Answer generation.} 
Then, we prompt the LLM $f$ to generate the answer $\hat{a}^f$ for each question-passage pair $\{q,p\}$, by feeding them as model input (prompts can be found in Appendix \ref{sec:appendix_rag_prompt}):
\begin{equation}
    \hat{a}^f = f(q,p).
\end{equation}

\item \textbf{Judgment annotation.} 
After that, we annotate $j$ for each tuple of $\{q,p,\hat{a}^f\}$.
As mentioned above, this judgment label indicates whether the RAG answer could be kept depending on confidence assessment.
To align with the supporting degree of given knowledge, we measure whether a sample is answerable approximately according to the correctness of the RAG answer $\hat{a}^f$, i.e.,
\begin{equation}
j =
    \begin{cases}
    1, & \text{if } \hat{a}^f \text{ is correct,} \\
    0, & \text{otherwise.}
    \end{cases}
\end{equation}
The correctness can be measured based on the ground-truth answer $a$, through Exact Match (EM) score, F1 score and so on.
Details can refer to Appendix~\ref{sec:appendix_data}.

\end{itemize}

Finally, we obtain two RC-RAG datasets, i.e., RC-TQ and RC-NQ.
The dataset statistics is shown in Table \ref{tab:statistics}.

\begin{figure*}[t]
    \centering
    \includegraphics[width=1\textwidth]{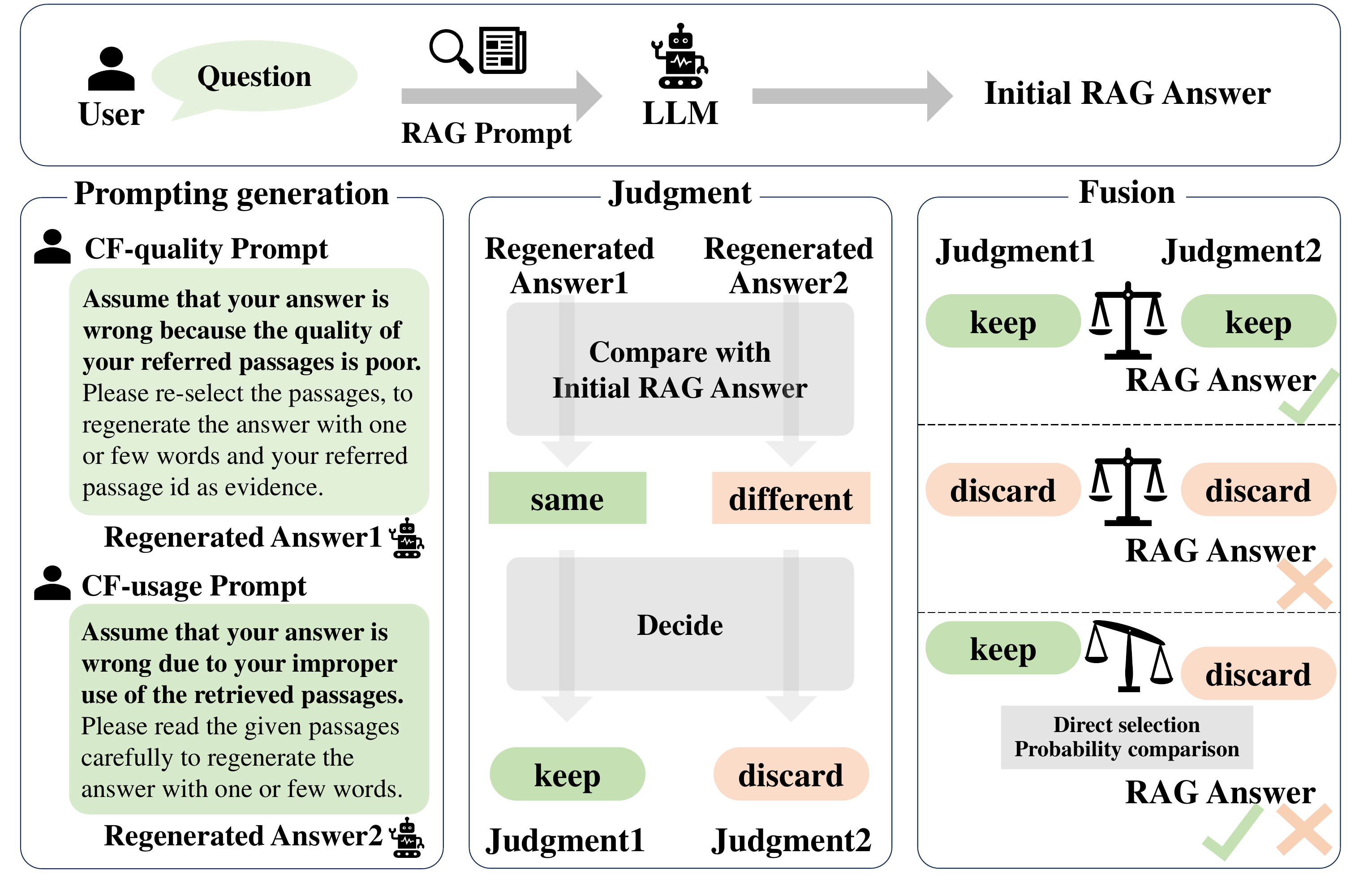}
    \vspace{-4mm}
    \caption{Overview of counterfactual prompting framework for RAG, in which the counterfactual (CF) prompts challenge the initial RAG answer in terms of the quality or  usage of retrieved results.
    The final judgment result is derived from both aspects.
    Details refer to Sec. \ref{sec:method}.}
    \label{fig:method}
    \vspace{-3mm}
\end{figure*}

\heading{Evaluation}
\label{sec:metric}
According to our RC-RAG benchmark, the samples could be divided into two cases, which are answerable (A) and unanswerable (U). 
Answerable ones refer to the samples whose RAG answer is correct, while unanswerable ones are the opposite.
At the same time, there are two prediction results for RAG answers based on the designed judgment strategy, i.e., keep (K) and discard (D).

By combining above situations, the output of RAG would fall into one of the four folds, i.e., AK, AD, UD or UK, as shown in the Table \ref{tab:risk}.
Specifically, AK/UK denotes the answerable/unanswerable samples with answers kept, while AD/UD denotes the answerable/unanswerable samples with answers discarded.
Noted that samples answered wrongly are labeled as unanswerable ones based on our annotation, thus there is no case of keeping the wrong answer in the answerable samples.

Among these four folds, we further analyze which one causes the real risk in RAG.
\begin{enumerate*}[label=(\roman*)]
\item It is intuitive that the AK and UD folds pose no risk, as the judgment results are consistent with the labels. 
\item For AD fold, although the judgment result is inconsistent with the label, it poses no real risk since the user' behaviour may not be influenced when the RAG provides a null answer.
\item Thus, only the UK fold exists risk, where the RAG sample is unanswerable but its answer is not discarded.
\end{enumerate*}

For evaluation, we propose four risk-aware evaluation metrics from various aspects, i.e., \textit{risk}, \textit{carefulness}, \textit{alignment} and \textit{coverage}.
\begin{table}[t]
    \centering
   \renewcommand{\arraystretch}{1}
    \setlength\tabcolsep{7.5pt}
    \begin{tabular}{ccc}
        \toprule
        & \multicolumn{2}{c}{\textbf{Judgment result}}\\
        \cmidrule(r){2-3}
        & Keep (K) & Discard(D)\\
        \midrule
        Answerable (A) &AK&AD\\
        Unanswerable (U) &UK&UD\\
        \bottomrule
    \end{tabular}
    \caption{Categorization of the RAG output.}
    \label{tab:risk}
    \vspace{-4mm}
\end{table}

\begin{itemize}[leftmargin=*]
\item \textbf{Risk} (\%) measures the percentage of risky cases (UK) among kept samples, i.e.,
$$risk = \frac{|\text{UK}|}{|\text{AK}|+|\text{UK}|},$$
where $||$ represents the number of samples.

\item \textbf{Carefulness} (\%) representing the percentage of incorrect samples being discarded, which is recall for unanswerable samples, i.e.,
$$ carefulness = \frac{|\text{UD}|}{|\text{UK}|+|\text{UD}|}.$$

\item \textbf{Alignment} (\%) represents the percentage of samples where the judgment results are consistent with the labels, i.e.,
$$alignment = \frac{|\text{AK}|+|\text{UD}|}{|\text{AK}|+|\text{AD}|+|\text{UK}|+|\text{UD}|}.$$

\item \textbf{Coverage} (\%) measures the percentage of samples to be kept, i.e.,
$$coverage = \frac{|\text{AK}|+|\text{UK}|}{|\text{AK}|+|\text{AD}|+|\text{UK}|+|\text{UD}|}.$$

Note that a lower \textit{risk} score is better, whereas higher scores are better for the other metrics.

\end{itemize}

\section{Counterfactual prompting framework}
\label{sec:method}

\heading{Overview}
To achieve risk control for RAG, we propose a novel counterfactual (CF) prompting framework that assesses predictive uncertainty of RAG.
The overview is illustrated in Figure \ref{fig:method}, consisting of a prompting generation module, a judgment module, and a fusion module:
\begin{enumerate*}[label=(\roman*)]
\item a prompting generation module, which utilizes counterfactual thinking to induce answer regeneration effected by two changing factors;
\item a judgment module, which makes judgment based on uncertainty assessment by analyzing the effect of each changing factor on their answer; and
\item a fusion module for the final judgment result.
\end{enumerate*}

\heading{Prompting generation module}
In this work, we assume that two latent factors can affect RAG uncertainty, i.e., the quality and the usage of retrieved results.
Thus, we argue about each of them and ask for answer regeneration, respectively.
Specifically, we implement each prompt as shown in Figure \ref{fig:method}, where CF-quality prompt challenges the poor quality of retrieved results and CF-usage prompt challenges the improper usage.
By imagining two scenarios that challenge each factor, the model adjusts the way it gets answers depending on its confidence level.

\heading{Judgment module}
This module decides whether to keep or discard the answer according to uncertainty assessment for both scenarios.
Specifically, we compare the regenerated answer with the initial RAG answer to analyze the effect of changing factors.
There are two kinds of comparison results, i.e., same or different.
Accordingly, the decision is made as follow:
\begin{enumerate*}[label=(\roman*)]
\item \textit{Keep}: Answer remaining the same indicates that the RAG answer is of relatively high confidence, which can be kept;
\item \textit{Discard}: Answer changing indicates that the RAG answer is uncertain, which should be discarded.
\end{enumerate*}

To reduce the likelihood of overestimating confidence, the prompting generation and judgment modules can be executed iteratively for $N$ rounds to validate the decision for each scenario. A decision is made as \textit{keep} only if the answer remains consistent across all $N$ rounds. To balance computational efficiency, we have set $N$ to 1.

\heading{Fusion module}
We aggregate above judgment results as below.
\begin{enumerate*}[label=(\roman*)]
\item If the two judgment results are consistent (both are \textit{keep} or \textit{discard}), we follow this judgment directly;
\item Otherwise (one is \textit{keep}, the other is \textit{discard}), make the final judgment according to following prompts-based strategies (prompts can be found in Appendix \ref{sec:appendix_fusion_prompt}):
\end{enumerate*}
\begin{itemize}[leftmargin=*]

\item \textbf{Direct selection:}
We prompt the LLM to make a final decision, by telling it potential reasons resulting in wrong answers chosen from [\textit{improper use} or \textit{poor quality}] of retrieval results, according to the scenario in which the \textit{discard} judgment was made in the previous judgment module.

\item \textbf{Probability comparison:}
We prompt the LLM to derive the probabilities of their respective judgments under two scenarios. By comparing the two probabilities, we select the judgment with the higher probability as the final judgment results.

\end{itemize}
 
After fusion, we change the judgment result of a special case from \textit{keep} to \textit{discard}: when the result is \textit{keep} and the RAG output is "unknown". In this case, keeping the result of "unknown" is equivalent to discarding.

More details and the complete form of all the prompts can refer to Appendix \ref{sec:appendix_details}, \ref{sec:appendix_prompt}.

\section{Experiment settings}
\label{sec:experiment_settings}

\heading{Baselines}
We compare our proposed CF prompting framework with three prompt-based baselines:
\begin{enumerate*}[label=(\roman*)]

\item \textbf{If-or-Else (IoE) prompting framework} \cite{li2024confidence}, facilitating self-corrections based on LLMs' confidence. 
To adapt to the RC-RAG, we classify the case of answer correction as discard.

\item \textbf{Calibration-based framework} \cite{tian2023just}, verbalizing confidence scores after obtaining answers, with a threshold set over verbalized scores. If the score is below the threshold, then choose to discard the output.

\item \textbf{Priori judgement framework} \cite{ren2023investigating}, perceiving the factual knowledge boundary by self-judgment in the normal or RAG setting, which discards an answer by saying "unknown".

\end{enumerate*}
More information about the baselines and their prompts can be found in Appendix \ref{sec:appendix_baseline},\ref{sec:appendix_prompt_baseline}.

\begin{table*}[t]
  \centering
  \renewcommand{\arraystretch}{1}
  \setlength\tabcolsep{0.68pt}
    \begin{tabular}{cccccccccc}
    \toprule
    \multirow{2}{*}{\textbf{Backbone}}&\multirow{2}{*}{\textbf{Method}}&\multicolumn{4}{c}{\textbf{RC-TQ}} & \multicolumn{4}{c}{\textbf{RC-NQ}}\\
    \cmidrule(r){3-6}
    \cmidrule(r){7-10}
    && risk$\downarrow$ & carefulness$\uparrow$ & alignment$\uparrow$ & coverage$\uparrow$ & risk$\downarrow$ & carefulness$\uparrow$ & alignment$\uparrow$ & coverage$\uparrow$ \\
     \midrule

     \multirow{4}{*}{\textbf{Mistral}}&IoE 
     & 24.88 & 20.97& 74.41 & \underline{91.06}
     & 45.59 & 20.22 & 56.93 & \underline{86.29} \\
     &Calibration 
     & 24.79 & 20.92 & \underline{74.80} & \textbf{91.47}
     & 45.65 & 17.36 & 57.06 & \textbf{89.25} \\
     &Priori 
     & \underline{21.95} & \underline{33.87} & \textbf{77.14} & 86.38
     & \underline{42.61} & \underline{28.60} & \underline{61.52} & 82.63 \\
     &Ours 
     & \textbf{19.00} & \textbf{52.87} & 72.78 & 71.14
     & \textbf{38.22} & \textbf{52.98} & \textbf{63.60} & 60.66 \\
     \midrule
     
     \multirow{4}{*}{\textbf{ChatGPT}}&IoE 
     & 21.59 & 33.53 & 78.88 & \textbf{88.34} 
     & 41.79 & 31.14 & 64.29 & \textbf{83.38}\\
     &Calibration 
     & 19.71 &42.51	&\underline{79.45}	&\underline{83.75}
      & 40.97 & 35.34 & 64.96 & \underline{79.78}\\
     &Priori 
     & \underline{16.23}	 &	\underline{57.30} &	\textbf{79.68} &	75.49 
     & \textbf{34.72}  & \underline{55.23} & \textbf{70.55} & 65.26\\
     &Ours 
     & \textbf{14.94} &	\textbf{65.37} &75.38 &	66.55 
     & \underline{35.22} & \textbf{62.86} & \underline{66.23} & 53.24\\
     
     \bottomrule

    \end{tabular}
  \caption{Main results of RC-RAG on the test set of two datasets and two LLMs with dense retriever. Best results in bold and second best in underline.}
  \label{tab:main_result}
\end{table*}

\begin{table}[t]
  \centering
  \renewcommand{\arraystretch}{1}
  \setlength\tabcolsep{0.1pt}
    \begin{tabular}{ccccc}
    \toprule
    \textbf{Method}&risk$\downarrow$ & carefulness$\uparrow$ & alignment$\uparrow$ & coverage$\uparrow$  \\
     \midrule
     \multicolumn{5}{c}{\textit{Sparse retrieval}} \\
     \hline
     IoE & 65.18 & 30.55 & 47.73 & 75.18\\
     Calibration & 65.10 & 28.98 & 47.31 & 76.98\\
     Priori & 60.43 & 43.15 & 56.70 & 66.37\\
     Ours & 56.30 & \textbf{65.80} & \textbf{65.15} & 42.85\\
     \hline
     \multicolumn{5}{c}{\textit{Dense retrieval}} \\
     \hline

     IoE 
     & 45.59 & 20.22 & 56.93 & 86.29 \\
     Calibration
     & 45.65 & 17.36 & 57.06 & \textbf{89.25} \\
     Priori
     & 42.61 & 28.60 & 61.52 & 82.63 \\
     Ours
     & \textbf{38.22} & 52.98 & 63.60 & 60.66 \\

     \bottomrule

    \end{tabular}
  \caption{Results of RC-RAG on the RC-NQ test set and Mistral with sparse retriever and dense retriever.}
  \vspace{-4mm}
  \label{tab:sparse_result}
\end{table}

\heading{Backbones}
We leverage two LLMs as backbones: Mistral \cite{jiang2024mixtral} and ChatGPT \cite{roumeliotis2023chatgpt}, which belong to open-source models and black-box models respectively.
Note that these methods are general and can be extended to other LLMs.

\heading{Implementation details} 
For LLMs, we call OpenAI’s API\footnote{platform.openai.com} to achieve ChatGPT (version gpt-3.5-turbo-0301), while we choose Mistral-7b\footnote{huggingface.co/mistralai/Mistral-7B-Instruct-v0.2} to implement Mistral.
The max sequence length of LLM output is set to 256, and the temperature is set to 0.
All the others are set as default.
For the retrieved results, we conduct dense retrieval and sparse retrieval following \citet{ren2023investigating}, and provide top-3 passages for each question following \citet{wang2023self}.
Most of the experimental results of our method use the direct selection fusion strategy, unless otherwise stated. 
More details refer to Appendix \ref{sec:appendix_details}. 
According to the analysis on the iteration number, as shown in Figure \ref{fig:iteration} in Appendix \ref{sec:appendix_details}, we report all results derived from a single run.

\section{Experiment results}
\label{sec:result}
We aim to answer six research questions: 
\textbf{(RQ1)} Does our CF prompting framework efficiently control the risk of RAG compared with the baseline methods?
\textbf{(RQ2)} Does the ability of LLMs affect the effectiveness of RC-RAG?
\textbf{(RQ3)} Does the difficulty of QA task affect the ability of RC-RAG?
\textbf{(RQ4)} Does the quality of retrieval results affect the effectiveness of RC-RAG? 
\textbf{(RQ5)} How does two CF-prompts affect the effectiveness of RC-RAG respectively? 
\textbf{(RQ6)} Are our risk control framework interpretable?

\subsection{Main results}
As shown in Table \ref{tab:main_result}, we present the performance of different RC-RAG methods on two datasets.
We have the following observations for \textbf{RQ1-3}.

\heading{Our approach effectively reduces risk and maintains carefulness compared to baselines}
Baselines without a clear indication of the possible source of error struggle to reject uncertain RAG answers:
(1) IOE has the worst rejection performance. For example, when using ChatGPT as a generator, it had the highest risk score and the lowest carefulness score on both datasets.
This suggests that directly judging confidence in the answer is difficult to overcome the LLM's overconfidence problem in the RAG setting, due to reliance on retrieved results.
(2) The calibration-based approach also suffers from overconfidence, resulting in the worst scores for risk and carefulness on both datasets when using Mistral as a generator.
This shows that LLMs tend to output high confidence scores in the RAG setting without considering the potential misdirection of retrieved results.
(3) The priori approach performs better on both metrics, particularly on the risk score of RC-NQ, achieving the lowest risk score of 34.72\% with ChatGPT.
This improvement is due to the prompt's mention of "based on the given information," leading the LLM to focus more on the quality of the retrieved results.

Our method outperforms the baselines in 3 out of the 4 settings (2 models and 2 datasets), achieving an average reduction of 2.88\% on risk scores and an average improvement of 14.77\% on carefulness scores.
The results show that uncertainty prediction based on retrieval results explicitly can effectively help risk control.
At the same time, alignment scores are not significantly inferior, especially on the RC-NQ dataset.
However, as trade-off, the performance of coverage is inferior to the baseline method.
It demonstrates how to balance risk control with coverage remains a difficult task.

\heading{Risk control ability is dependent on the LLM ability}
We compare the performance of RC-RAG when using different LLMs as generators.
We find that risk control works better with ChatGPT than with Mistral. 
Benchmark statistics (Table \ref{tab:statistics}) show that Mistral outperforms ChatGPT on both datasets, particularly on RC-NQ.
This indicates that risk control is more effective with weaker LLMs, underscoring the necessity of risk control methods. The underlying reason is that more capable models are more confident in both their internal knowledge and retrieved results. Consequently, Mistral achieves higher coverage scores, demonstrating that stronger LLMs tend to retain answers, which is consistent with the reasons for the above results.

\begin{table}[t]
  \centering
  \renewcommand{\arraystretch}{1}
  \setlength\tabcolsep{0.1pt}
    \begin{tabular}{ccccc}
    \toprule
    \textbf{Method}& risk$\downarrow$ & carefulness$\uparrow$ & alignment$\uparrow$ & coverage$\uparrow$\\
     \midrule

     Ours & 13.56 & 75.75 & 76.00 & 59.00\\
     \midrule
     CF-usage& 18.31 & 60.61 & 78.00 & 71.00\\
     CF-quality& 10.48 & 83.33 & 74.50 & 52.50\\

     \bottomrule

    \end{tabular}
  \caption{Ablation study results of RC-RAG on the subset of RC-TQ test set and ChatGPT with dense retriever.}
  \label{tab:ablation}
  \vspace{-3mm}
\end{table}

\begin{table*}[h]
  \small
  \renewcommand{\arraystretch}{1} 
  \setlength\tabcolsep{5pt}
    \centering

    \begin{tabularx}{\textwidth}{X}
      \toprule
      \textbf{Question:} What was the purpose of a revival meeting during the second great awakening?
      \\
      \textbf{Ground-truth answer:} Evangelizing.\\
      \textbf{Passage:} 
      [Passage-1]: Title: Second Great Awakening Content: a sense of personal salvation. Upon their return home, most converts joined or created small local churches, which grew rapidly. Calvinists emphasized the inability of men to save themselves. The only way to be saved was by God's electing grace. The Revival of 1800 in Logan County, Kentucky, began as a traditional Presbyterian sacramental occasion. 
      ... [Passage-2]: ... [Passage-3]: ...\\
      \midrule
      \textbf{RAG answer:} The purpose of a revival meeting during the Second Great Awakening was to inspire individuals to seek personal salvation and join or create small local churches. Evidence: Passage-1\\
      \midrule
      \textbf{IOE:} The purpose of a revival meeting during the Second Great Awakening was to inspire individuals to seek personal salvation and join or create small local churches. Evidence: Passage-2, Passage-1 (\textcolor{blue}{Keep})\\
      \textbf{Calibration:} Probability: 0.95 (\textcolor{blue}{Keep})\\
      \textbf{Priori:} To inspire personal salvation and encourage individuals to turn to Christ.(\textcolor{blue}{Keep})\\
      \hline
      \textbf{CF-usage:} The purpose of a revival meeting during the Second Great Awakening was to inspire individuals to seek personal salvation and join or create small local churches, as well as to provide an opportunity for settlers to encounter organized religion and socialize with others. (\textcolor{blue}{Keep})\\
      \textbf{CF-quality:} Sorry, there is no information provided in the given passages about the purpose of a revival meeting during the Second Great Awakening. (\textcolor{red}{Discard})\\
      \textbf{Fusion:} \textcolor{red}{Discard}\\

      \bottomrule
    \end{tabularx}

    \caption{An example (No.135) from the RC-NQ test data, to analyze the generated answers and judgments of different risk control method for RAG.
    We mark the correct judgments in red and wrong ones in blue.}
    \label{tab:case}
    \vspace{-2mm}
  \end{table*}

\heading{Task difficulty has limited influence on risk control ability}
We compared the effect of RAG risk control methods on different tasks.
According to the risk and alignment scores, we find that the risk control methods perform worse in RC-NQ than in RC-TQ.
The statistics of the benchmark (Table \ref{tab:statistics}) show that RC-NQ is significantly more difficult than RC-TQ, as both ChatGPT and Mistral have a lower percentage of answerable samples on the RC-NQ dataset.
We find that the more difficult the task to answer, the more difficult the risk control.
For coverage scores, the performance in RC-NQ is also weaker.
However, the performance in terms of carefulness scores was largely flat.
The conclusion drawn from the above phenomenon is that \textit{the difficulty of the task has a limited effect on the ability of the risk control method to accurately identify unanswerable samples}.
As the proportion of samples that cannot be answered is larger in tasks with higher difficulty, the proportion of samples (UK) that cannot be answered but are retained will also be larger, and the risk and coverage scores will be correspondingly increased.

\subsection{Impact of retriever}
\label{sec:retriever}

To answer \textbf{RQ4}, we compared the performance of risk control of RAG with different retrievers.
Results are shown in Table \ref{tab:sparse_result}, conducted on the RC-NQ test set using Mistral as a generator.

By comparing the results using different retrievers, we observe that the risk control method is more cautious with the sparse retriever in terms of carefulness. 
However, the sparse retriever results in significantly more unanswerable samples than the dense retriever (Table \ref{tab:complete_statistics} in Appendix~\ref{sec:appendix_data}), leading to a higher risk score.
Additionally, the experimental results show that our method outperforms all baselines using both retrievers in terms of risk, carefulness, and alignment.

\subsection{Analysis of CF prompt and fusion strategy}
\label{sec:ablation}

To answer \textbf{RQ5}, we conduct ablation study to investigate the effects of the two CF prompts separately.
The experiment was conducted on a subset of the RC-TQ test set using ChatGPT as a generator.
We used CF-quality and CF-usage separately in prompting generation module, followed by the judgement module.
The experimental results are shown in the Table \ref{tab:ablation}, from which we have the following observations.

\heading{Only CF-usage prompting}
The effect of risk control decreases while the coverage score increases, indicating that the model tends to stick to its answer when confronted with challenge about the usage of retrieved results.
This shows that the model is confident about the usage of retrieved results, which is essentially the internal knowledge of the LLMs, consistent with their characteristics of overconfidence.

\heading{Only CF-quality prompting}
In contrast to the above, the risk score decreases significantly, indicating that the model tends to modify its answers when confronted with challenge about the quality of retrieved results.
This shows that the model is sensitive to the challenge of the quality of retrieved results, which belongs to external knowledge, and the model itself does not have the ability to judge the quality of external knowledge.

\heading{Fusion strategy} 
The comparison results using two different fusion strategies are shown in Table \ref{tab:fusion_result} in Appendix \ref{sec:appendix_fusion_comparison}. 
Our complete approach with fusion module can effectively balance the two situations, considering both risk and coverage.
Specifically, the direct fusion strategy can identify the unanswerable samples more effectively.

\subsection{Case study}

To answer \textbf{RQ6}, we conduct a case study to illustrate the working mechanism of our method, based on ChatGPT augmented with dense retrieval.

As shown in Table \ref{tab:case}, the RAG answer and its referred passages inaccurately address the question, yet no baseline methods reject to answer.
Our approach, while unable to detect errors when the usage of retrieved passages is challenged, recognizes their quality limitation and abstains from providing an answer.

\section{Conclusion}

In this work, we propose a counterfactual prompting framework for assessing the uncertainty of RAG results, based on the quality of the retrieved results and the manner in which they are used.
We construct a benchmark and design risk-related evaluation metrics. Experimental results with two LLMs on two datasets show that our method can effectively reject unanswerable samples and has a certain interpretability.
In the future, we will explore other factors that may affect predictive uncertainty in RAG, such as conflicts between internal and external knowledge. Additionally, we will attempt to design objective functions based on risk-related metrics to guide the joint learning of the risk control framework and the RAG model.


\section*{Limitations}

Firstly, the two latent factors influencing RAG's confidence are human-defined, which may not encompass the full spectrum of risk sources.
Future work could explore more diverse factors identified by LLMs, combined with statistic analysis.

Methodologically, our prompting generation approach is computationally intensive. Further exploration is needed to develop more efficient prompting strategies. 
The judgment module currently struggles with long answers, which requires a more sophisticated matching function.
Additionally, the current fusion strategy is heuristic. Future enhancements could include semantic information to better integrate the two judgments.

Furthermore, we have focused solely on risk control in a zero-shot scenario. 
How to improve RAG answers in this scenario deserves further investigation.
Also, designing objective functions based on risk-related metrics for joint training with the RAG framework could be explored, aiming for a balanced trade-off between risk control and response quality.

\section*{Ethics statement}

We have emphasized ethical considerations at every stage to ensure the responsible application of AI technologies. 
This work does not utilize personally identifiable information or require manually annotated datasets. 
Our methods are transparent, and we have made our data and code public to facilitate reproducibility and further research.

\section*{Acknowledgments}

This work was funded by the National Natural Science Foundation of China (NSFC) under Grants No. 62372431 and 62472408, the Strategic Priority Research Program of the CAS under Grants No. XDB0680102, XDB0680301, the National Key Research and Development Program of China under Grants No. 2023YFA1011602 and 2021QY1701, the Youth Innovation Promotion Association CAS under Grants No. 2021100, the Lenovo-CAS Joint Lab Youth Scientist Project, and the project under Grants No. JCKY2022130C039.

\bibliography{custom}

\clearpage
\appendix

\section{Details about annotation}
\label{sec:appendix_data}

At the judgment annotation stage, we define the following criteria: 
Given the ground-truth answer $a$ and the RAG answer $\hat{a}$, if $EM(a,\hat{a})=1, F1>\tau, RougeL>\tau$, or the $a$ appears in $\hat{a}$, the RAG answer can be judged as correct, and the sample can be annotated as answerable.
We set $\tau=0.7$.

\begin{table}[h]
    \centering
   \renewcommand{\arraystretch}{1}
    \setlength\tabcolsep{2.5pt}
    \begin{tabular}{ccccc}
        \toprule
         & \multicolumn{2}{c}{\textbf{Sparse-RC-NQ}}   & \multicolumn{2}{c}{\textbf{Dense-RC-NQ}}  \\
        \cmidrule(r){2-3}
        \cmidrule(r){4-5}
        & S-NQ-A& S-NQ-U& D-NQ-A& D-NQ-U \\
        \midrule
        Mistral & 1063 & 2547 & 1830 & 1780\\
        
        \bottomrule
    \end{tabular}
    \caption{The statistics of the full test sets of RC-NQ and annotated results of answerable (A) and unanswerable (U) samples, utilizing Mistral as the generator with both sparse and dense retrievers.}
    \label{tab:complete_statistics}
    \vspace{-4mm}
\end{table}

\section{Implementation details}
\label{sec:appendix_details}

\heading{Details of judgment module}
The criteria for determining that the answers remain unchanged are consistent with the criteria for matching the answers in the judgment annotation stage (Appendix \ref{sec:appendix_data}).
If the regenerated answer matches the RAG answer, it can be judged as \textit{same} and thus \textit{keep}.

\begin{figure}[h]
    \centering
    \includegraphics[width=1\linewidth]{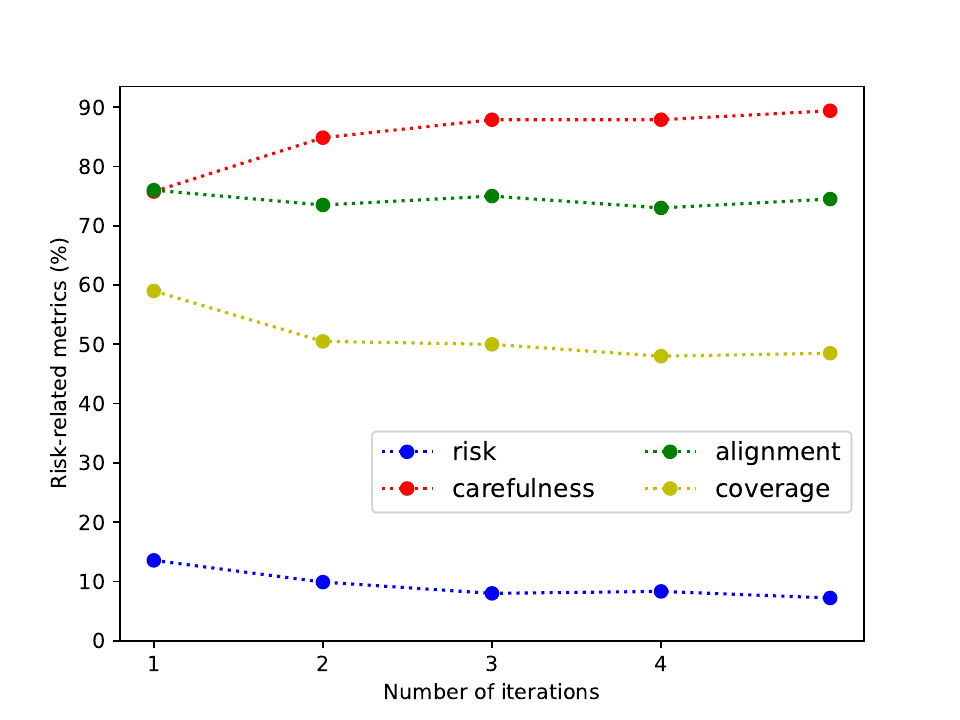}
    \caption{The change of risk-related metrics with the increase of iteration number.}
    \label{fig:iteration}
\end{figure}

\heading{Details of iterative process}
The number of our iterative process $N$ is chosen from [1,2,3,4,5].
Specifically, we explored the performance of risk control when the number of iterations increased from 1 to 5, and the experimental setting was the same as Sec. \ref{sec:ablation}.
The results are shown in the figure \ref{fig:iteration}, we can find that:
with the increase of iterations, risk and coverage score showed a downward trend, carefulness score increased, while the alignment index was basically flat. 
In order to save the computational cost, we chose the number of iterations to be 1 to carry out the rest of our experiments.

\section{Prompt for CF prompting framework}
\label{sec:appendix_prompt}

\subsection{Prompt for basic RAG setting}
\label{sec:appendix_rag_prompt}
\heading{RAG prompt}
\textit{Answer the following question based on the given passages with one or few words. Provide your evidence between two \#\# symbols at the end of your response, either the passage id or your internal knowledge. For example, provide "Answer: apple. Evidence: \#\# Passage-0, Passage-1 \#\#." if you are referring to Passage-0 and Passage-1 to obtain the answer "apple". If there is no information in the passages, explain the answer by yourself.}

\textit{Question: \{question\}} 

\textit{Passages: \{passage\}}

\subsection{Prompt for prompting generation}

\heading{CF-quality prompt}
\textit{Assume that your answer is wrong because the quality of your referred passages is poor. Please re-select the passages, to regenerate the answer with one or few words and your referred passage id as evidence.}

\heading{CF-usage prompt}
\textit{Assume that your answer is wrong due to your improper use of the retrieved passages. Please read the given passages carefully to regenerate the answer with one or few words.}

\begin{table*}[t]
  \centering
  \renewcommand{\arraystretch}{1}
  \setlength\tabcolsep{1pt}
    \begin{tabular}{cccccccccc}
    \toprule
    \multirow{2}{*}{\textbf{Backbone}}&\multirow{2}{*}{\textbf{Method}}&\multicolumn{4}{c}{\textbf{RC-TQ}} & \multicolumn{4}{c}{\textbf{RC-NQ}}\\
    \cmidrule(r){3-6}
    \cmidrule(r){7-10}
    && risk$\downarrow$ & carefulness$\uparrow$ & alignment$\uparrow$ & coverage$\uparrow$ & risk$\downarrow$ & carefulness$\uparrow$ & alignment$\uparrow$ & coverage$\uparrow$ \\
     \midrule

     \multirow{2}{*}{\textbf{Mistral}}
     &Ours$_{pro}$
     & 21.23 & 43.37& 72.68 & \textbf{76.48}
     & 41.72 & 44.49 & 60.17 & \textbf{65.60} \\

     &Ours$_{dir}$ 
     & \textbf{19.00} & \textbf{52.87} & \textbf{72.78} & 71.14
     & \textbf{38.22} & \textbf{52.98} & \textbf{63.60} & 60.66 \\
     \hline
     
     \multirow{2}{*}{\textbf{ChatGPT}}
     &Ours$_{pro}$
     & 16.30 & 59.96 & \textbf{79.26} & \textbf{70.55} 
     & 36.24 & 57.87 & \textbf{66.65} & \textbf{58.70}\\

     &Ours$_{dir}$ 
     & \textbf{14.94} &	\textbf{65.37} &75.38 &	66.55 
     & \textbf{35.22} & \textbf{62.86} & 66.23 & 53.24\\
     
     \bottomrule

    \end{tabular}
  \caption{Comparison results of our methods using two different fusion strategies, on the test set of two datasets and two LLMs with dense retriever. The subscripts $_{dir}$ and $_{pro}$ represent the use of direct selection strategy and probability comparison strategy, respectively.}
  \label{tab:fusion_result}
\end{table*}

\subsection{Prompt for fusion}
\label{sec:appendix_fusion_prompt}
\heading{Direct selection prompt}

\begin{itemize}[leftmargin=*]

\item \textit{Your answer is likely to be wrong because of the poor quality of retrieval passages, please choose to keep or discard this output. Generate \$\$ keep \$\$ if you choose to keep this answer, otherwise, generate \$\$ discard \$\$.}

\item \textit{Your answer is likely to be wrong because of the improper use of retrieval passages, please choose to keep or discard this output. Generate \$\$ keep \$\$ if you choose to keep this answer, otherwise, generate \$\$ discard \$\$.}

\end{itemize}

\heading{Probability comparison prompt}
\textit{Provide the probability that your regenerated answer is correct. Give ONLY the probability, no other words or explanation.}

\textit{For example:}

\textit{Probability: <the probability between 0.0 and 1.0 that your specific answer is correct, without any extra commentary whatsoever; just the probability!>}

\section{Baselines}
\label{sec:appendix_baseline}

Among the three baseline methods, IoE and calibration-based framework are post-processing methods, while priori judgment framework is a pre-processing method.

\textbf{IoE method} was originally used for answer correction, requiring the model to update the answer of low confidence. If the model updates the answer, guide it to choose a final answer.
Based on the matching results between the final answer and the RAG answer, we decide whether to keep or discard the RAG answer.

\textbf{Calibration-based framework} requires a threshold to discard answers.
We set the threshold as 0.6 based on the experimental results.

\textbf{Priori judgment framework} requires prompt input only once, which explicitly mentions "given information" and "internal knowledge" in its prompt.

\section{Prompt for baselines}
\label{sec:appendix_prompt_baseline}

\heading{IOE prompt}

\begin{itemize}[leftmargin=*]

\item \textit{If you are very confident about your answer, maintain your answer. Otherwise, update your answer.}

\item \textit{You give two different answers in previous responses. Check the problem and your answers again, and give the best answer.}
\end{itemize}

\heading{Calibration prompt}
\textit{Provide the probability that your answer is correct. Give ONLY the probability, no other words or explanation.}

\textit{For example:}

\textit{Probability: <the probability between 0.0 and 1.0 that your specific answer is correct, without any extra commentary whatsoever; just the probability!>}

\heading{Priori prompt}
\textit{Given the following information: }

\textit{\{passage\}}

\textit{Can you answer the following question based on the given information or your internal knowledge, if yes, you should give a short answer with one or few words, if no, you should answer "Unknown".}

\textit{Question: \{question\}}

\section{Analysis of fusion strategies}
\label{sec:appendix_fusion_comparison}

We show the comparison results of our methods using two different fusion strategies in Table \ref{tab:fusion_result}.

\section{AI Tool Usage Instructions}
We utilized ChatGPT to assist in refining the expressions and wording of the paper.

\end{document}